\documentclass[10pt,journal,compsoc]{IEEEtran}

\ifCLASSOPTIONcompsoc
  \usepackage[nocompress]{cite}
\else
  \usepackage{cite}
\fi

\usepackage{gensymb}
\usepackage{mathtools}
\usepackage{rotating}

\usepackage{booktabs}
\usepackage{subcaption}
\usepackage{algorithm}
\usepackage{algorithmic}
\usepackage{enumitem}
\usepackage{multirow}
\usepackage{wrapfig}
\usepackage{epsfig}
\usepackage{graphicx}
\usepackage{color}
\usepackage[table]{xcolor}
\usepackage{epsfig}
\usepackage{graphicx}
\usepackage{amsmath}
\usepackage{amssymb,siunitx,subcaption}
\usepackage{footmisc}
\usepackage[pagebackref=true,breaklinks=true,letterpaper=true,colorlinks,bookmarks=false]{hyperref}

\usepackage{pifont}
\usepackage[capitalize]{cleveref}
\crefname{section}{Sec.}{Secs.}
\Crefname{section}{Section}{Sections}
\Crefname{table}{Table}{Tables}
\crefname{table}{Tab.}{Tabs.}

\usepackage{booktabs}
\usepackage{cuted}
\usepackage{capt-of}
\usepackage{cuted}
\usepackage{multirow}
\usepackage{capt-of,etoolbox}	
\makeatletter

\begin{document}

\title{Progressive Learning of 3D Reconstruction Network from 2D GAN Data}

\author{
Aysegul Dundar, Jun Gao,  Andrew Tao, Bryan Catanzaro

\IEEEcompsocitemizethanks{
\IEEEcompsocthanksitem  A. Dundar, J. Gao, A. Tao,  B. Catanzaro are  with NVIDIA, CA, USA.
\IEEEcompsocthanksitem A. Dundar is with Department of Computer Science, Bilkent University, Ankara, Turkey
\IEEEcompsocthanksitem J. Gao is with Department of Computer Science, University of Toronto, Canada.
}
}


\IEEEtitleabstractindextext{
\begin{abstract}
This paper presents a method to reconstruct high-quality textured 3D models from single images.
Current methods rely on  datasets with expensive annotations; multi-view images and their camera parameters.
Our method relies on GAN generated multi-view image datasets which have a negligible annotation cost.
However, they are not strictly multi-view consistent and sometimes GANs output distorted images. This results in degraded reconstruction qualities.
In this work, to overcome these limitations of generated datasets, we have two main contributions which lead us to achieve state-of-the-art results on challenging objects: 1) A robust multi-stage learning scheme that gradually relies more on the models own predictions when calculating losses, 2) A novel adversarial learning pipeline with online pseudo-ground truth generations to achieve fine details.
Our work provides a bridge from 2D supervisions of GAN models to 3D reconstruction models and removes the expensive annotation efforts.
We show significant improvements over previous methods whether they were  trained on GAN generated multi-view images or on real images with expensive annotations. Please visit our web-page for 3D visuals: \href{https://research.nvidia.com/labs/adlr/progressive-3d-learning}{https://research.nvidia.com/labs/adlr/progressive-3d-learning}.
\end{abstract}
}

\maketitle
\IEEEdisplaynontitleabstractindextext
\IEEEpeerreviewmaketitle

\section{Introduction}

GAN based models achieve realistic image synthesis on various objects ~\cite{karras2019style, karras2020analyzing, yu2021dual} and find applications in image editing, conditional image generation \cite{park2019semantic,dundar2020panoptic,liu2022partial}, and video generation \cite{wang2021one} tasks.
They are also found to be useful for dataset generations with automatic part segmentation annotations \cite{zhang2021datasetgan, tritrong2021repurposing}.
There is further interest to deploy this technology for gaming, robotics, architectural designs, and AR/VR applications.
However, such applications also require contollability  on the viewpoint requiring generation in 3D representations.
On the other hand, the realism of 3D image generation and reconstruction results are not on par with the GAN generated 2D images \cite{bhattad2021view, monnier2022share, yu2021pixelnerf, chan2022efficient, gao2022get3d}.
In this work, we are interested in closing this gap.

To reconstruct high quality  3D models, current state-of-the-art (SOTA) methods rely on 3D annotations. Such data is expensive to collect, requires special hardware and is usually collected in constrained lab environments. 
Due to the difficulty in collecting such annotations, efforts have been limited to few objects such as faces \cite{gecer2019ganfit} and human bodies \cite{zhang2019predicting, lattas2020avatarme}.
A cheaper  alternative is finely curated multi-view datasets.
They can be collected with a camera without expensive hardware requirements.
However, they are still difficult to annotate for their camera parameters. 
For that reason, synthetic images are used instead of real images to train 3D reconstruction models \cite{chang2015shapenet, choy20163d, chen2019learning, yu2021pixelnerf}.
While these models learn to reconstruct synthetic objects, 
they fall short in their ability to recover the 3D properties of real images due to the domain gap between synthetic and real images.

\begin{figure*}
    \centering
    \includegraphics[width=1.0\linewidth]{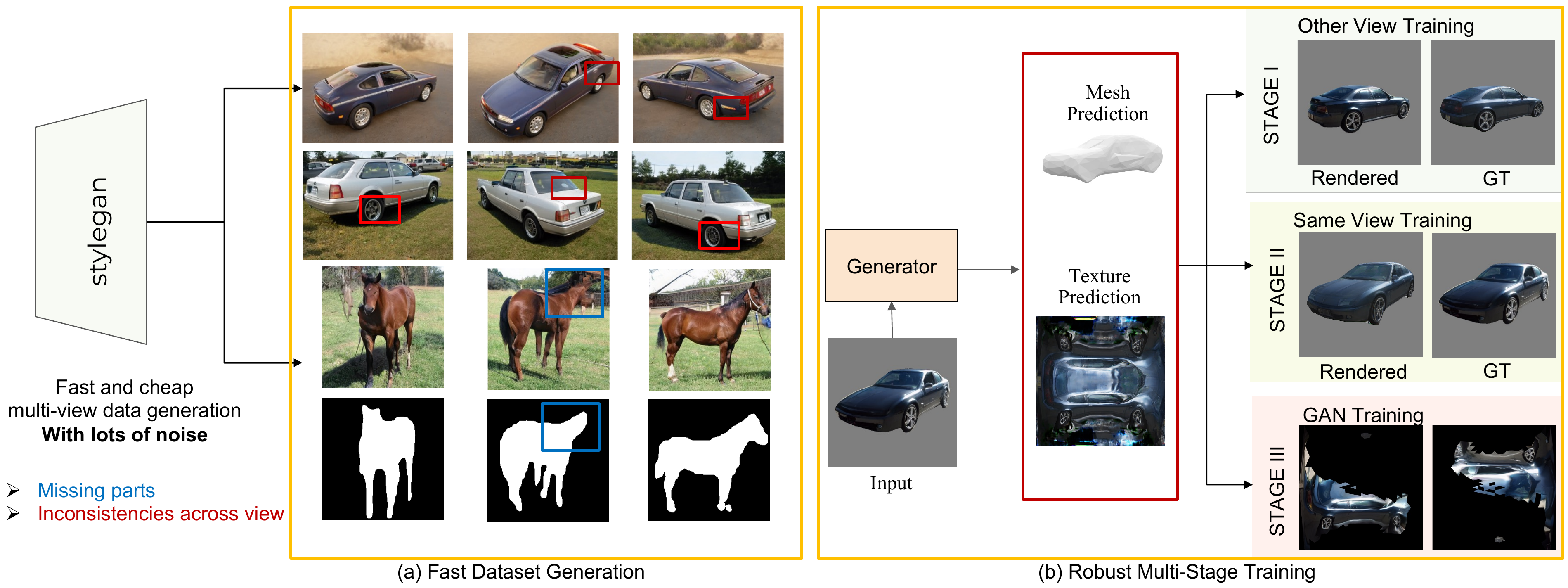}
    \caption{Overview of the dataset generation (a) and multi-stage training scheme of the reconstruction network (b). The generator network takes input image and outputs mesh and texture predictions. In the first stage, the output is rendered from another view than the input image and losses are calculated on this novel view. This way model is not effected by the missing parts in the images and also the unrealistic segmentation maps resulted from these images.
    In the second stage, additional reconstruction loss is added from the same view.
    The rendered and ground-truth images are masked based on the silhouette predictions of the model.
    Lastly, to achieve sharp and realistic predictions, we add adversarial training on the third stage. 
    GAN training pipeline is given in Fig. \ref{fig:gan-overview}.}
    \label{fig:motoverview}
\end{figure*}

Single view image collections are also explored to learn 3D reconstruction models \cite{kanazawa2018learning, bhattad2021view}. 
However, with single view images during trainings, models receive limited supervision; only to the visible parts. Various constraints and regularizers are proposed to obtain plausible results such as losses which limit the deformation from mean template \cite{kanazawa2018learning, chen2019learning}, rotation and swap consistency losses \cite{bhattad2021view, monnier2022share} and semantic consistency \cite{li2020self} constraints. 
Still results are not realistic.
Another way to use single view image collections to learn 3D reconstruction models is to train a GAN model from them and generate multi-view datasets with the trained GAN models \cite{zhang2020image,liu20222d}.
This approach becomes possible because recent generative models of images, especially StyleGANs, are shown to learn an implicit 3D representation with latent codes that can be manipulated to change the viewpoint of a scene~\cite{karras2019style}.
The latent codes of the StyleGAN are controlled to generate consistent objects from different view points.
Few selected viewpoints are labeled for camera parameters which only takes a minute to annotate and unlimited number of samples can be generated \cite{zhang2020image} for those viewpoints.
However, one issue of these datasets is that they are not perfect especially in presenting the realistic details across views.
This is because StyleGAN does not have strict disentanglement of shape, texture, and camera parameters.
Therefore, one cannot change the camera parameters while preserving the identity strictly.
Another issue is the distorted image generations, sometimes appear as missing parts in objects (cf. Fig. \ref{fig:motoverview}).

\begin{figure*}
    \centering
   \includegraphics[width=\textwidth]{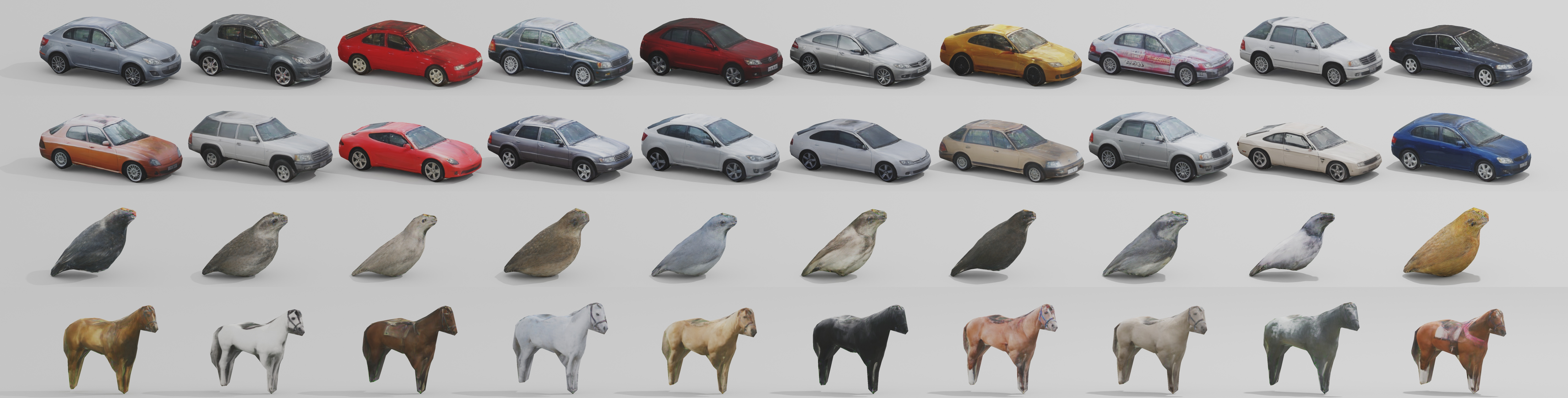}
    \caption{Given a single 2D image input, our method outputs high quality textured 3D models. We achieve these results by learning from StyleGAN generated datasets via a robust multi-stage training scheme and a novel adversarial learning pipeline.}
    \label{fig:teaser}
\end{figure*}

In this work, our goal is to learn  accurate 3D reconstruction  models from GAN generated multi-view images.
As our first contribution, we propose a framework that is robust to the noise in the training data.
We achieve this with a  multi-stage learning scheme that gradually relies more on the models own predictions when calculating losses. 
Secondly, we propose a novel adversarial learning pipeline with online pseudo-ground truth generation to train a discriminator. 
With the discriminator, our model learns to output fine details. 
Our model shows significant improvements over previous methods whether they were  trained on GAN generated multi-view images or on real images with expensive data collections/annotations pipelines.
In summary, our main contributions are:
\begin{itemize}[leftmargin=*]
    \item A robust multi-stage learning scheme that relies more on the model's predictions at each step. Our model is not affected by missing parts in the images  and inconsistencies across views.
    \item A novel adversarial learning pipeline to increase the realism of textured 3D predictions. 
    We generate pseudo-ground truth during training and employ a multi-view conditional discriminator for learning to generate fine details.
    \item High-fidelity textured 3D model synthesis both qualitatively and quantitatively on three challenging objects. 
    Examples are shown in Fig. \ref{fig:teaser}. 
\end{itemize}

\section{Related Work}

Style-based GAN models~\cite{karras2019style, karras2020analyzing} achieve high quality synthesis of various objects which are quite indistinguishable from real image and are shown to learn an implicit 3D knowledge of objects without a supervision.
One can control the viewpoint of the synthesized object by its latent codes.
This makes pretrained GANs a promising technology for controllable generation \cite{wang2022high,alaluf2022hyperstyle,pehlivan2022styleres}.
However, in these models, the disentanglement of 3D shape and appearance is not strict and therefore the appearance of objects change as the viewpoint is manipulated.
Recently, 3D-aware generative models are proposed 
 with impressive results but they either do not guarantee strict 3D consistency \cite{nguyen2019hologan, niemeyer2021giraffe, chan2021pi, gu2021stylenerf} or computationally expensive \cite{chan2022efficient} and overall not on par with 2D StyleGAN results \cite{gao2022get3d}. 
 We are interested in single-view image inference so our work is more related to image inversion methods that project images into these 3D-aware GAN's latent space \cite{yin20223d, ko20233d, liu20223d}.
 Even though significant progress is achieved for image inversion \cite{yin20223d, ko20233d, liu20223d}, these methods require run-time optimization and suffer from lower quality novel view predictions.

There have been many works that learn textured 3D mesh models  from images  with differentiable renderers~\cite{loper2014opendr, kato2018neural, liu2019soft, chen2019learning, ravi2020accelerating}.
Deep neural networks are coupled with the renderers and trained to predict 3D mesh representations and texture maps of input images via reconstruction losses~\cite{kanazawa2018learning, chen2019learning, goel2020shape, li2020self, henderson2020leveraging}.
However, inferring these 3D attributes from  single 2D images is inherently ill-posed problem  given that the invisible mesh and texture predictions receive no gradients during training ~\cite{kanazawa2018learning, chen2019learning, goel2020shape, li2020self, henderson2020leveraging}.
These algorithms that learn from single-view images output results that look unrealistic especially  when viewed from a different point.

Multi-view image datasets provide a solution for the limited supervision problem of single-view image datasets.
However, due to the expensive annotation of 
multi-view image datasets for their 2D keypoints or camera pose, they are small in scale.
There have been methods that use sequence of images to optimize a mesh and texture model \cite{dundar2022fine}.
However, they learn a new network for each sequence.
Recently, these sequence of image datasets have also been tremendously explored with a method called Neural Radiance Fields (Nerf) to explore implicit geometry \cite{mildenhall2020nerf, martin2021nerf, wang2021nerf}.
These models overfit to a sequence and can not be used for single image inference.
PixelNerf \cite{yu2021pixelnerf} is an extension of these models that achieves single image inference, however, as we show in our experiments, the results are not good.

Another promising direction with Nerf-based models is optimization of 3D representations with well-trained diffusion models.
These models can stylize meshes or generate 3D geometry representations from scratch with given text prompts \cite{metzer2022latent, lin2022magic3d, michel2022text2mesh, poole2022dreamfusion}.
However, these models require run time-optimization and control on the generation is limited. 
In our work, we are interested in a different application, single view image reconstruction where the generation is conditioned on an input image.

In our work, we are interested in mesh representations due its efficiency in rendering.
To infer mesh representations, multi-view datasets are also shown to be beneficial based on the experiments with synthetic datasets~\cite{chang2015shapenet, tulsiani2017multi, tulsiani2018multi}.
However, those results do not translate well to real image inferences because of the domain gap between synthetic and real images.
To generate a realistic multi-view dataset with cheap labor cost, Zhang et. al.~\cite{zhang2020image} use a generative adversarial network by controlling the latent codes and generate coarsely consistent objects from different view points.
We also use these datasets but achieve significantly better results than the state-of-the-art and Zhang et. al.~\cite{zhang2020image} thanks to the robust learning scheme (which also allows us to remove regularizers that limit the deformations) and the adversarial learning pipeline.


\section{Method}

In Section \ref{sec:mot}, we describe the motivation of our approach. 
The multi-stage training scheme and adversarial learning pipeline are presented in Section \ref{sec:multi} and \ref{sec:adv}, respectively.

\subsection{Motivation}
\label{sec:mot}

Differentiable rendering enable training neural networks to perform 3D inference such as predicting 3D mesh geometry and textures from images \cite{chen2019learning}.
However, they require  multi-view images, camera parameters, and object silhouettes to achieve high performance models.
Such data is expensive to obtain.
StyleGAN generated datasets remove the expensive labeling effort via the latent codes that control the camera viewpoints.
When few viewpoints are selected and annotated, multi-view images can be generated in infinite numbers for those viewpoints.
The annotations require 1 minute \cite{zhang2020image} no matter how many images are generated because they are all aligned across different examples.
As for the segmentation masks the renderers utilize during training, they can be obtained by off-the-shelf instance segmentation models \cite{he2017mask}. 
However, learning a high performing model from these datasets remains a challenge since generated images suffer from precise multi-view consistency as shown in Fig. \ref{fig:motoverview} by red rectangles.
Additionally, some examples have missing parts as shown in Fig. \ref{fig:motoverview} by blue rectangles, the head of the horse is not generated in good quality which also transfers to instance segmentation mask (fourth-row).
In this work, we address these challenges by proposing a robust multi-stage training pipeline and an adversarial learning set-up.

\begin{figure*}
    \centering
    \includegraphics[width=1.0\linewidth]{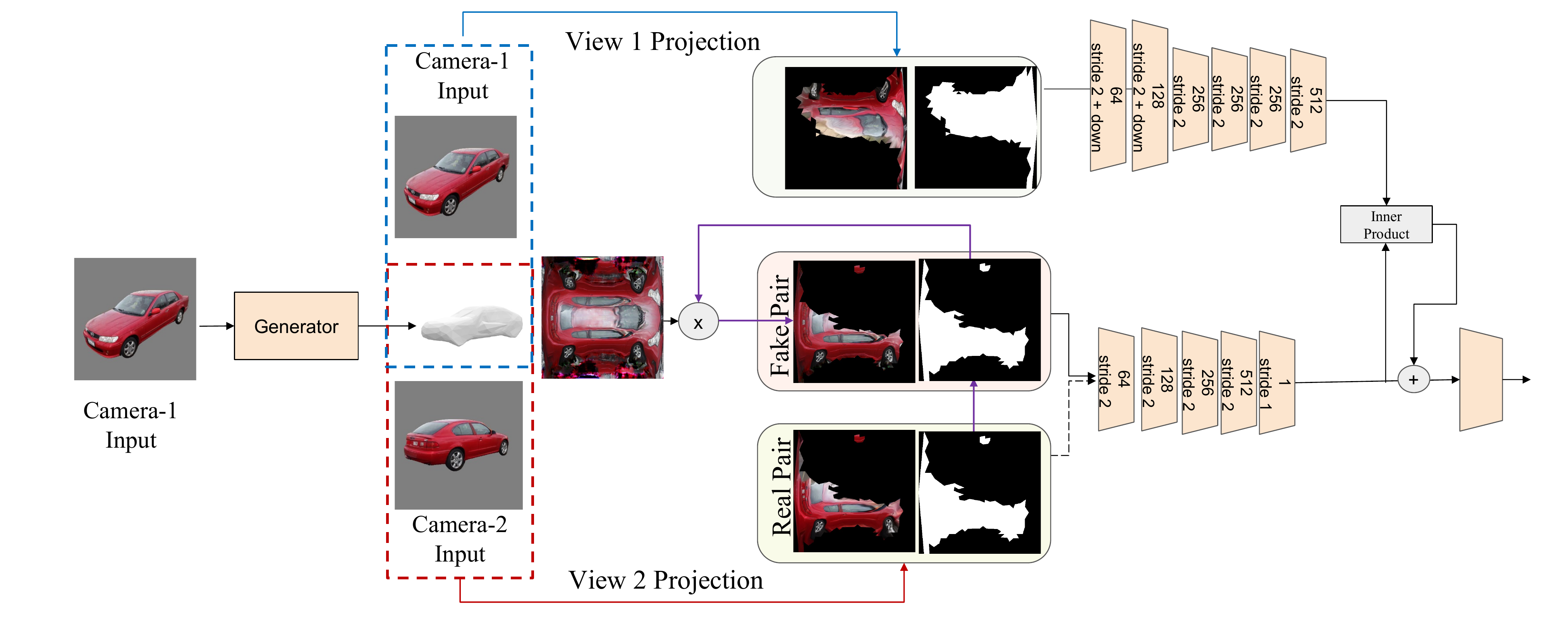}
    \caption{Gan training pipeline. Discriminator is trained with fake and real pairs and a conditioning pair.
    First, texture and mesh predictions are outputted by the generator.
    With the estimated mesh and given camera parameters of a second view image, texture and visibility maps are projected from the second view image.
    For fake pair, estimated texture is partially erased by the visibility map.
    Additionally, conditioning input pair is obtained by projecting from the first view image with the estimated mesh predictions and given camera parameters of the first view.}
    \label{fig:gan-overview}
\end{figure*}

\subsection{Multi-stage Training Pipeline}
\label{sec:multi}

We train the generator that outputs mesh and texture predictions with multi-stage training pipeline to be robust to the errors in annotations and multi-view inconsistencies. 
At each stage of our pipeline, results improve progressively.

\textbf{First Stage.} In the first stage, our model outputs 3D mesh and texture predictions from an input image that is from camera view-1, $I^g_{v1}$. 
We render the image from these 3D predictions with the target view, view-2, and output the image $I^r_{v2}$. We calculate the losses on these novel predictions.
The target view is randomly selected among the sequence.
The motivation of the first stage is to capture reliable 3D mesh predictions and reasonable texture estimations.
At this stage, we do not expect a high quality texture estimation given the inconsistencies across views. 
Our experiments show that when the network is guided with the losses from the same view as input ($I^r_{v1}$ vs. $I^g_{v1}$),  the errors in missing parts and so the errors in segmentation mask annotations propagate to the mesh predictions.
This instabilizes the training even when the model is trained with multi-view consistency, e.g. objectives calculated from both $I^r_{v1}$ and $I^r_{v2}$.
Therefore, in the first stage, we learn to reconstruct an object from the image of the object from a different view. 
This way model does not overfit to the errors of the given view since it receives feedback from a novel view.
Note that novel view may and does also have errors but since a novel view is randomly sampled from a sequence and errors are not consistent among the views, 
the model outputs the most plausible 3D model to minimize losses in a sense similar to majority voting. 

The training losses at this stage is calculated as follows. 
We use perceptual image reconstruction loss between the ground-truth image of the target view ($I^g_{v2}$) and the rendered image view ($I^r_{v2}$). We mask the images with ground-truth silhouette (mask) predictions, $S^g_{v2}$. This way reconstruction loss is only calculated on the object.
As a reconstruction loss, we use perceptual losses from Alexnet ($\Phi$) at different feature layers ($j$) between these images from the loss objective as given in Eq. \ref{eq:nv-percep}.


\begin{equation}
\label{eq:nv-percep}
\small
    \mathcal{L}_{p-nv} = ||\Phi_{j}(I^g_{v2}*S^g_{v2}) - \Phi_j(I^r_{v2}*S^g_{v2}) ||_2
\end{equation}
For shapes, we use an IoU loss between the silhouette rendered ($S^r_{v2}$) and the silhouette ($S^g_{v2}$) of the input object.
\vspace{-1pt}
\begin{equation}
\label{eq:sil}
\vspace{-1pt}
\small
\mathcal{L}_{sil} = 1 - \frac{||S^g_{v2}\odot S^r_{v2} ||_1}{||S^g_{v2}+ S^r_{v2} - S^g_{v2}\odot S^r_{v2} ||} 
\end{equation}

Similar to~\cite{chen2019learning, liu2019soft}, we also regularize predicted mesh using a  laplacian loss ($\mathcal{L}_{lap}$) constraining neighboring mesh triangles to have similar normals.  Following are our base losses:
\vspace{-2pt}
\begin{equation}
\begin{multlined}
\small
\mathcal{L}_{first} = 
\lambda_{pn} \mathcal{L}_{p-nv} + \lambda_s \mathcal{L}_{sil}  +  \lambda_{lap} \mathcal{L}_{lap}   
\end{multlined}
\end{equation}

The model outputs reliable 3D mesh predictions since it gets feedback from different views.
Note that, we do not use many regularizes such as mean template and penalizing deformation vertices as previous works ~\cite{chen2019learning, bhattad2021view, zhang2020image} and still achieve a stable training with the first stage objectives.


\textbf{Second Stage.} In the second stage, we rely on our 3D mesh predictions and introduce additional losses between $I^r_{v1}$ vs. $I^g_{v1}$, same view as the input image.
In 3D inference predictions, we expect the model to output predictions that faithfully match with object for the input view.
Therefore, in the second stage, we add
additional reconstruction losses from the input
view.
However, we do not add silhouette loss for the input view because there are noise in the segmentation masks. 
Furthermore, for the reconstruction loss, we do not mask the input image and rendered output with the ground-truth mask since it is noisy.
We rely on the 3D prediction of our model and mask the reconstruction loss based on the projected mesh prediction.

The rendered and ground-truth images are masked based on the silhouette predictions of the model and  perceptual loss is calculated as given in Eq.  \ref{eq:sv_perc}. This way, we rely on the first stage training for the mesh prediction and learn improved textures with the same view training for the visible parts. Mesh predictions can still improve via the reconstruction losses since they still receive a feedback via image reconstruction, but we do not guide it directly with the silhouettes.

\begin{equation}
\label{eq:sv_perc}
\small
    \mathcal{L}_{p-sv} = ||\Phi_{j}(I^g_{v1}*S^r_{v1}) - \Phi_j(I^r_{v1}*S^r_{v1})||
\end{equation}

We use the following loss in additional iterations:

\begin{equation}
\begin{multlined}
\small
\mathcal{L}_{second} =  \mathcal{L}_{first}   
+ \lambda_{ps} \mathcal{L}_{p-sv} 
\end{multlined}
\end{equation}

The second stage training starts after first stage training converges.
That is because we rely on the models mesh predictions in the newly introduces losses.
Results improve significantly but the results are not sharp as the training data.

\textbf{Third Stage.} 
After learning reliable mesh representation and high quality texture predictions, we use generative learning pipeline to improve realism of our predictions.
In this stage, we rely on our predictions to generate pseudo ground-truths to enable adversarial learning which is explained in the next section.

\subsection{Adversarial Learning}
\label{sec:adv}
Training the model with an adversarial loss applied on the rendered images do not improve the results due to the shortcomings of renderers \cite{pavllo2020convolutional}.
Therefore, we convert texture learning into 2D image synthesis task.
Texture learning in UV space is previously explored with successful results 
\cite{gecer2019ganfit, pavllo2020convolutional, dundar2022fine}.
However, our set-up is different as we learn a single view image inference, the texture projection and pseudo ground-truth generations are online in our training, we do not learn a GAN trained from scratch for texture generation rather we tune our 3D reconstruction network, and we propose a multi-view training in our discriminator.
While previous methods use different networks for texture projection and texture generation, we achieve both with the same architecture.
This enables us to improve the generator further and achieve state-of-the-art results.

As shown in Fig. \ref{fig:gan-overview}, during our training, we obtain projected texture maps for the input view ($v1$) and a different view ($v2$).  We obtain those by first predicting 3D meshes from an input view ($v1$) via our generator.
The input images are projected onto the UV map of the predicted mesh template based on the camera parameters of each image via an inverse rendering.
In this process, mesh predictions are transformed onto 2D screen by projection with camera parameters.
Then transformed mesh coordinates and UV map coordinates are used in reverse way and real images are projected onto UV map with the renderer.
Visibility masks are also obtained in this set-up.
With this set-up, we obtain a real partial texture (from $v2$) and a conditioning texture map (from $v1$) to train our discriminator.
In this set-up, it is important for the mesh predictions to be accurate for the correct inverse rendering. That is way we leave the GAN training to the third stage.

\textbf{Discriminator.}
Finally, to provide adversarial feedback, we train a conditional discriminator. The discriminator is conditioned on the partial texture view 1. Partial texture view 2 is a real example and the generated texture is a fake one.
For the fake example, we mask the generated texture with the visibility mask from real example to prevent distribution mismatch.

In traditional image-to-image translation algorithms conditional input and target images are concatenated and fed to the discriminators. 
However, in our case, the images are partially missing and aligning them in input via concatenation do not provide useful signals.
Instead, we use a projection based discriminator \cite{miyato2018cgans} where we process the conditioned input via convolutional layers and global pooling until the spatial dimension decreases to $1\times1$.
Again the reason to decrease the dimension is because the input is partially missing, therefore we want a full receptive field of the input image while conditioning on the patches.

We dot product the conditional input that is embedded and the discriminator outputs.
This score is added to the final discriminator score.
With the multi-view conditioning, the discriminator does not only consider if the patch is realistic but also if the predicted texture is consistent with its input pair.
The GAN training is especially important in our set-up since we do not have consistent multi-view images.
The overall objective for the third stage includes following min-max optimization:
%
\begin{align}
    \min_{\theta_g}\max_{\theta_d}~ \mathcal{L}_{second} + \lambda_{adv}\mathcal{L}_{adv}(\theta_g,\theta_d)
\end{align}
where  $\theta_d$ and $\theta_g$  refer to parameters of the discriminator and the generator, respectively.

\section{Experiments}

\textbf{Datasets.} First to generate datasets, we use three category-specific StyleGAN models, one representing a rigid object class, and two representing articulated class.
For car and horse dataset,  official models from StyleGAN2 (\cite{karras2019style}) repository are used. These models are trained on LSUN  Car dataset with 5.7M images and LSUN horse dataset  with 2M images \cite{yu2015lsun}. We also use a model trained on a bird class on NABirds dataset \cite{van2015building} with 48k images.
The StyleGAN generated images are aligned for few view-points and those views are annotated for one example which takes 1 minutes in total. Please refer to Zhang et. al. \cite{zhang2020image} for more details in the dataset generation pipeline.

\begin{figure}
    \centering
    \includegraphics[width=1.0\linewidth]{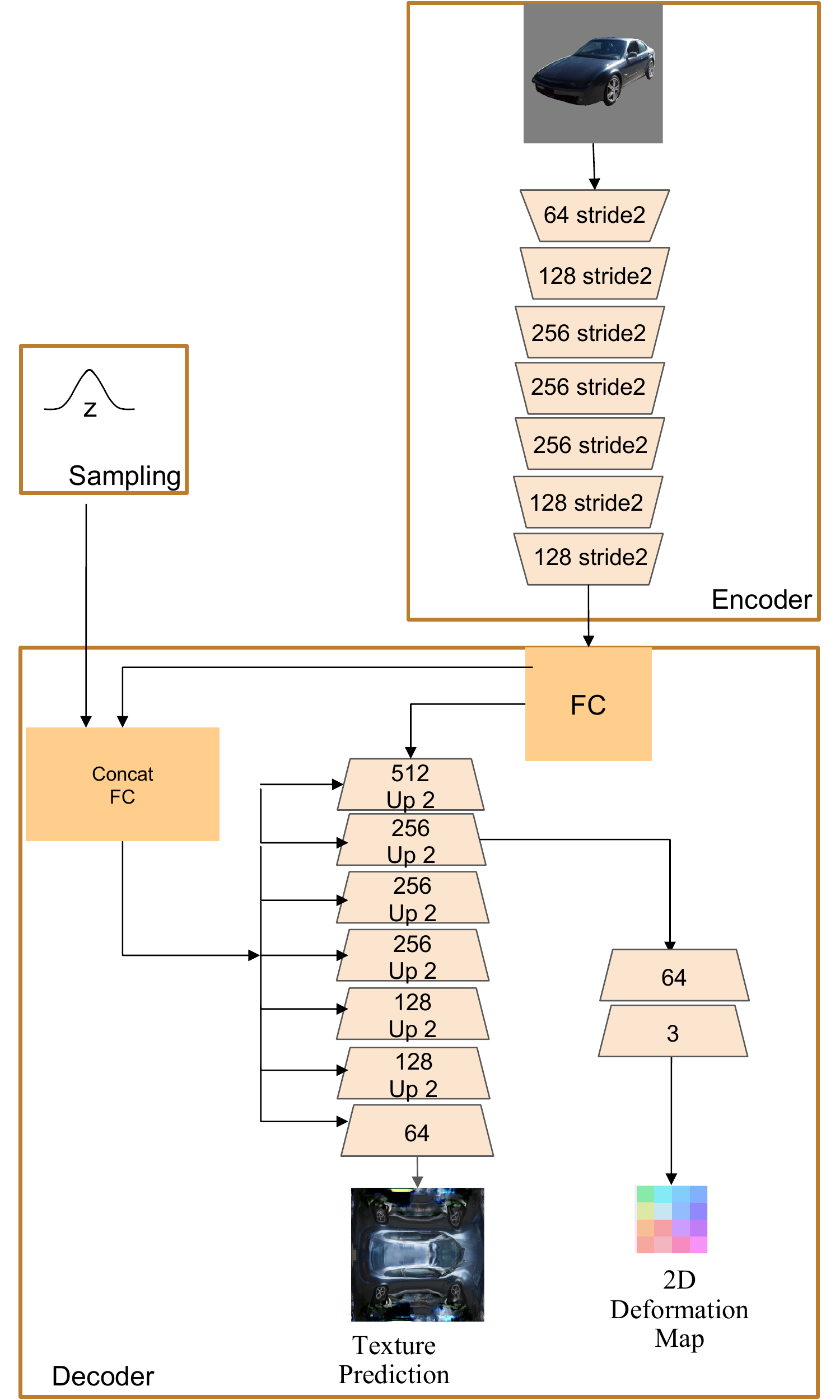}
    \caption{Encoder-decoder based generator architecturet that takes the input image and sampled $z$ to output texture and 2D deformation map predictions.  Each block represent a convolution layer with channel size. The encoder and decoder are connected with fully connected layers.}
    \label{fig:sup_gen}
\end{figure}

\noindent {\textbf{Architectural Details of Generator.} }
Our generator has an encoder-decoder architecture as shown in Fig. \ref{fig:sup_gen}. In the encoder, for predicting deformation and texture maps, the encoder receives $512\times512$ image.
It has 7 convolutional blocks with each convolution layer with $3\times3$ filters and stride of $2$.
The number of channels of the convolution layers are $(64, 128, 256, 256, 256, 128, 128)$.
There is a ReLU non-linearity after each convolution layer.
The encoder decreases the spatial size to $4\times4$.The encoded features ($128\times4\times4$) go through a fully connected layer and after reshaping, we output features with dimension of  $512\times8\times4$. Note that, we start with width of $8$ and height of $4$.
While generating texture and mesh predictions, we only predict half of the maps in the height since they are expected to be mostly symmetric in y axis.
Later, we flip the predictions and concatenate with itself  to expand the height.

The texture and mesh predictions go through two shared blocks of convolutional layers at first.
In the decoder, each block has two convolutional layers. There is an adaptive normalization and leaky ReLU after each convolutional layer. 
There is also a skip connection from input to the output at each block.
After each block, there is a  bilinear interpolation layer to upsample the feature maps at each layer.
The first two blocks has channels of $(512, 256)$ which bring the feature maps to a spatial dimension of $32\times16$.
After that mesh prediction branches out and there is another block with channel size of $64$ for the mesh prediction branch. After that there is a single convolutional layer with channel of $3$. The output  represents the deformations (x,y,z) coordinates.

\begin{figure}
    \centering
    \includegraphics[width=1\linewidth]{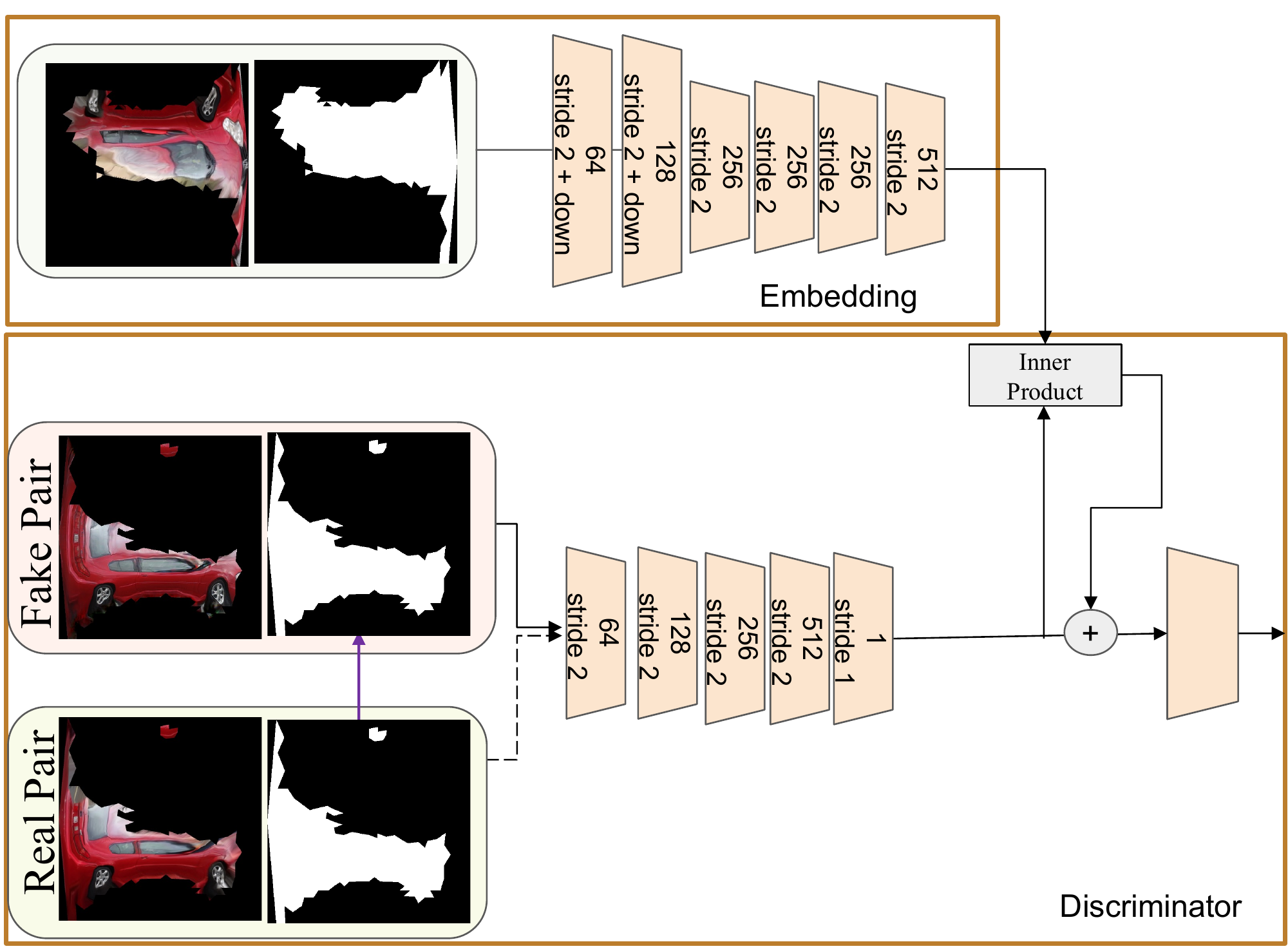}
    \caption{Discriminator architecture.  To provide adversarial feedback, we train a projection based conditional discriminator. }
    \label{fig:sup_discr}
\end{figure}

For texture prediction, there are 4 more convolution blocks with channels size of $(256, 256, 128, 128)$.
After these layers, the spatial resolution becomes $512\times256$.
A reflection symmetry is applied as we flip the texture predictions in y axis and concatenate it with the original texture predictions.
This results in spatial resolution of $512\times512$.
After that there is one more convolution block with channel size of $64$ and a final convolution layer which decreases the number of channels to $3$.
They represent (R,G,B) channels of the texture prediction.
The convolutional layers after symmetry relaxes the symmetry constrains as we do not expect a perfect symmetry in the texture.

Note that the mesh predictions are also estimated in a convolutional way as a UV deformation map
~\cite{pavllo2020convolutional,bhattad2021view,dundar2022fine}. 
Our deformation is a representation on the function of sphere directly with a fixed surface topology.
We sample from the deformation map for the corresponding vertex locations. 
We also apply symmetry on the predicted UV deformation.
We use DIB-R~\cite{chen2019learning} as our differentiable renderer. 
The renderer takes mesh and texture predictions and output images for target camera parameters.

We additionally sample a latent vector from normal distribution to provide a diversity in our predictions.
The sampled latent vector is concatenated with the encoded features via a linear layer.
The output is fed to the adaptive normalization layers in convolutional blocks.

\begin{figure*}
    \centering
    \includegraphics[width=1\linewidth]{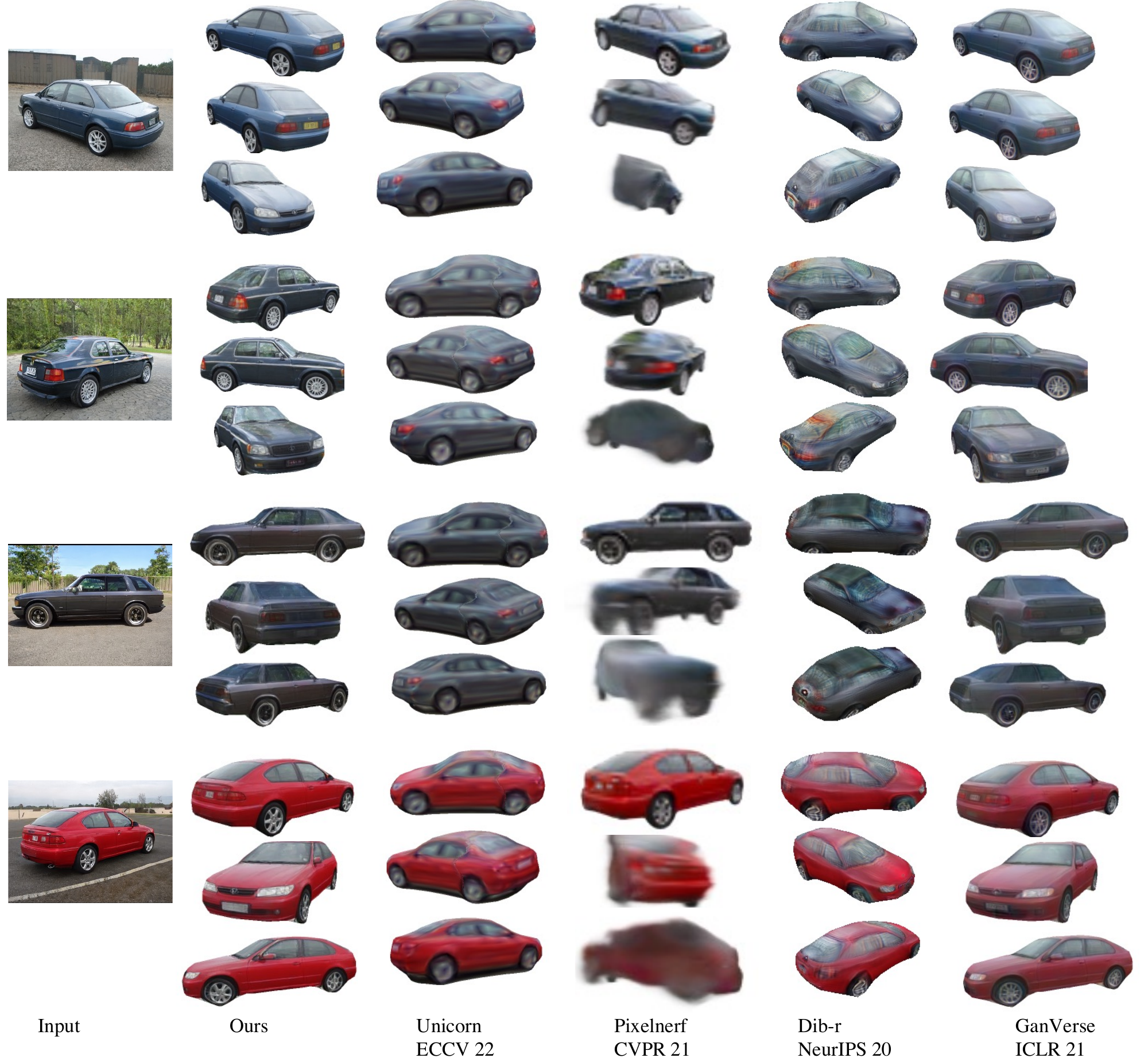}
    \caption{Given input images (1st column), we predict 3D shape, texture, and render them into the same viewpoint and novel view points for our model. We also show renderings of state-of-the-art models that are trained on real and synthetic images.
    Since, models use different camera parameters, we did not align the results.
    However, the results shown from similar viewpoints capture the behaviour of each model. Unicorn outputs similar 3D shape for different inputs. Pixelnerf achieves a high quality same view results but the results are poor from novel views. Dib-r also suffers from the same issue. 
    Ganverse does not output realistic details. 
    Our model achieves significantly better results than the previous works while being trained on a synthetic (StyleGAN generated) dataset.}
    \label{fig:results}
\end{figure*}

\noindent {\textbf{Architectural Details of Discriminator.} }
As for the architecture of the discriminator, we use a projection based discriminator \cite{miyato2018cgans}.
The discriminator takes the generated and real $3\times512\times512$ texture maps, and the pseudo ground-truth visibility mask as shown in Fig. \ref{fig:sup_discr}.
Generated textures are also multiplied with the masks to prevent a mismatch between the real-fake data distributions.
We concatenate the input with learnable positional embeddings on both scales which is omitted from the figure \cite{dundar2022fine}.
The discriminator adopts a multi-scale architecture with two scales one operates on $32\times32$ patches, the other $16\times16$.
We also have a conditioning pathway shown as an embedding network.
The embedding network process the conditioned input via convolutional layers and global pooling until the spatial dimension decreases to $1\times1$.
We dot product the conditional input that is embedded to $1\times1$ resolution and the discriminator outputs.
This score is added to the final discriminator score.
This happens for both scales of discriminators.

{\noindent \bf Training parameters} 
We train our model on 8 GPUs with batch size of 4 per GPU, for 100 epochs in total with learning rate of $1^{-4}$.
In our loss functions, we use $\lambda_{lap} = 0.5$, $\lambda_{p-sv} = 1$, $\lambda_{p-nv} = 1$, $\lambda_{adv} = 1$. 
The discriminators learning rate is set to $10^{-4}$.
We use Adam optimizer for updating both the 3d inference model and discriminator.



\begin{table*}[t]
\centering
\caption{
We report results for same view (input and target have the same view) and novel view (input and target have different view).
We provide FID scores, LPIPS, MSE, SSIM, and 2D mIOU accuracy predictions and GT.
We compare with GanVerse since they are trained on the same dataset. We also provide results of each stage showcasing improvements of progressive training.}
\begin{tabular}{|l|l|l|l|l|l|l|l|l|l|l|l|}
\hline
& & \multicolumn{5}{c|}{Same View} & \multicolumn{5}{c|}{Novel View}\\
\hline
 & \textbf{Method} &  FID$\Downarrow$ & LPIPS$\Downarrow$ & MSE$\Downarrow$ & SSIM$\Uparrow$ & IoU$\Uparrow$ & FID$\Downarrow$ & LPIPS $\Downarrow$ & MSE$\Downarrow$  & SSIM$\Uparrow$  &  IoU$\Uparrow$ \\
\hline
\multirow{4}{*}{\rotatebox[origin=c]{90}{Car}} & GanVerse \cite{zhang2020image} & 28.04 &0.1238  & 0.0060 &0.8683 &0.92 & 29.59 & 0.1333 & 0.0075 & 0.8599 & \textbf{0.93} \\
& Ours - Stage I & 10.75 & 0.1011 & 0.0060 & 0.8695 &  \textbf{0.94} & 11.98  & 0.1091 & 0.0074 & 0.8582 & 0.92\\
& Ours -Stage II & 6.24 &  0.0737 & \textbf{0.0039} &0.9027& \textbf{0.94} & 9.05 & 0.1012 & \textbf{0.0072} & \textbf{0.8651} & \textbf{0.93} \\
& Ours - Stage III & \textbf{4.56} & \textbf{0.0696} & 0.0040 & \textbf{0.9039} & \textbf{0.94} &  \textbf{6.92} &  \textbf{0.0965} & 0.0076 &  0.8632 & \textbf{0.93}\\
\hline
\multirow{4}{*}{\rotatebox[origin=c]{90}{Bird}} &
GanVerse \cite{zhang2020image} & 69.32 & 0.0742 &  0.0034 & 0.9230 & 0.82 & 63.89 & 0.0782 & 0.0037 & 0.9202 & 0.80 \\
& Ours - Stage I & 69.58 & 0.0763 & 0.0035 &  0.9222  &  0.82 & 63.76 & 0.0764  & 0.0036 & 0.9222 & 0.81\\
& Ours -Stage II & 64.67 & 0.0720 & 0.0030 & \textbf{0.9258} &  0.82 &  60.97 & 0.0749 & 0.0036 & \textbf{0.9226} & 0.81\\
& Ours - Stage III & \textbf{61.21} & \textbf{0.0689} & \textbf{0.0030} & 0.9231 &  \textbf{0.83} & \textbf{59.08} & \textbf{0.0712} & \textbf{0.0035} & 0.9201 &  \textbf{0.82}\\
\hline
\multirow{4}{*}{\rotatebox[origin=c]{90}{Horse}} &
GanVerse \cite{zhang2020image} & 62.38&  0.1272 & 0.0060 & 0.8852 & 0.78 &  83.82 & 0.1531 &  0.0100 & 0.8642 &  0.77 \\
& Ours - Stage I & 83.66 & 0.1395 & 0.0088 &  0.8727  &  0.78 & 76.64 & 0.1442  & 0.0100 & 0.8669 & \textbf{0.78}\\
& Ours - Stage II & 57.07 & 0.1037 & 0.0059 & 0.8971 & \textbf{0.79} &  67.77 & 0.1368 & 0.0101 & 0.8676 & \textbf{0.78}\\
& Ours - Stage III & \textbf{56.83} & \textbf{0.1024} & \textbf{0.0055} & \textbf{0.9017} &  \textbf{0.79} & \textbf{67.30} & \textbf{0.1367} & \textbf{0.0099} & \textbf{0.8684} &  \textbf{0.78}\\
\hline
\end{tabular}
\label{table:results_all}
\end{table*}

\begin{figure*}
    \centering
    \includegraphics[width=1\linewidth]{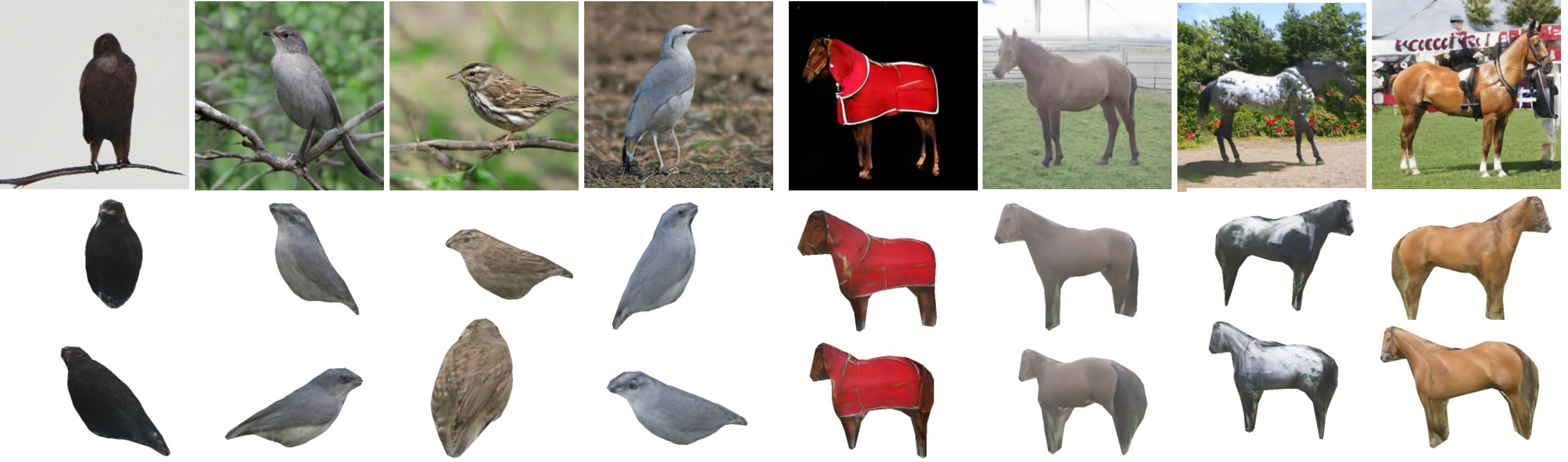}
    \caption{Qualitative results of our final model on Bird and Horse class.
     Given input images (1st row), we predict 3D shape, texture, and render them into the same view point and novel view points.}
    \label{fig:bird}
\end{figure*}

{\noindent \bf Evaluation.} We report various metrics on validation datasets. Since we have multi-view data, we report the scores for both same view and novel view. Same view results are obtained by rendering the predictions from the same view given as the input whereas for novel view, we render the predictions from a different camera view than the input. 
Same view looks at the fidelity to the given input whereas novel view measures if the model estimates the invisible texture and geometry which is a more difficult task.
For both views, we report Frechet Inception Distance (FID) metric \cite{heusel2017gans} which looks at the realism by comparing the target distribution and rendered images, Learned Perceptual Image Patch Similarity (LPIPS) \cite{zhang2018unreasonable} which compares the target and rendered output pairs at the feature level of a pretrained deep network, Structural
Similarity Index Measure (SSIM) and Mean Squared Error (MSE)  which compare the pairs in pixel-level similarity.  
We also report the intersection-over-union (IoU) between the target silhouette and projected silhouette of the predicted geometry.  


{\noindent \bf Results.} We provide quantitative results in Table~\ref{table:results_all}. 
We provide comparisons with GanVerse \cite{zhang2020image} which is trained on the same data as ours.
We observe quantitative improvements at each stage for all three classes as given in Table \ref{table:results_all} as well as large improvements over GanVerse model especially on FID metrics which measure the quality of generations.
We qualitatively compare with  other methods, since they all use different camera set-ups, it is not possible to do an accurate quantitative comparison.
However, on our qualitative result comparisons (Fig.~\ref{fig:results}), it is clear that our results achieve significantly better results.

In our qualitative comparisons, we compare with Unicorn \cite{monnier2022share} which learns 3D inference model in an unsupervised way on Pascal3D+ Car dataset \cite{xiang2014beyond}. The model only uses the bounding box annotation and is trained on 5000 training images. As can be seen from Fig.~\ref{fig:results}, impressive results are achieved given that the model is learned in an unsupervised way. On the other hand, the results lack details and diversity in the shapes. It is significantly worse than ours.
Second, we compare with Pixelnerf \cite{yu2021pixelnerf} which predicts a continuous neural scene representation conditioned on a single view image.
While neural radiance fields  \cite{mildenhall2020nerf} optimizes the representation to every scene independently, Pixelnerf trains across multiple scenes to learn a scene prior and is able to perform novel view synthesis given an input image. Pixelnerf is trained on a synthethic dataset, ShapeNet dataset \cite{chang2015shapenet}. It is also showcased on real image reconstruction for car classes. In our results, Pixelnerf is very good at predicting the same view but not as successful on the novel view predictions.
Next, we compare with Dib-r model \cite{chen2019learning} which is trained on Pascal3D+ Car dataset with ground-truth silhouette and camera parameters.
Dib-r outputs reasonable results on the same-view predictions of the input image. However, their texture and meshes do not generalize across views and results in unrealistic predictions from novel views even though the model is trained with expensive annotations.

\begin{figure}[t]
  \centering
  \vspace{-5pt}
  \includegraphics[width=\linewidth]{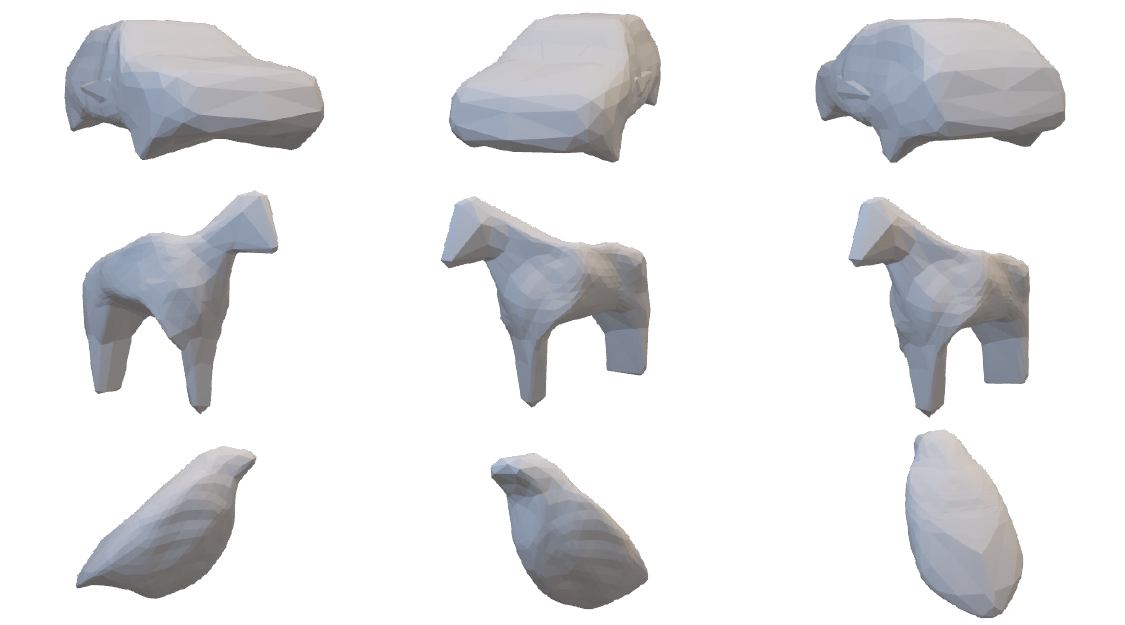}
  \vspace{-6pt}
  \caption{3D model predictions of our model.
  \vspace{-12pt}}
  \label{fig:reb-3d-mdoel}
 \end{figure}

Last, we compare with GanVerse model which is trained on the same StyleGAN generated dataset as our method.
As shown in Table \ref{table:results_all},
the quantitative results of GanVerse are even worse than our single-stage results. 
GanVerse model is trained with same view and other view reconstructions simultaneously. 
One difference is that,  GanVerse model learns a mean shape and additional deformation vertices for each image. The additional deformation is penalized  for each image for stable training. 
Since, we rely on other view training for mesh prediction, we do not put such constrain on the vertices. 
Their architecture is based on a U-Net and they do not employ a GAN training. Their results lack details and do not look as realistic  compared to our model.
Note that StyleGAN generated dataset also removes the expensive labeling effort and models that train on this dataset have the same motivation with the models that learn without annotations such as Unicorn.
Generating StyleGAN dataset with annotations requires 1 minute, whereas Pascal3D+ dataset requires 200h~350h work time for the annotations \cite{zhang2020image}.
Therefore, our comparisons provided in Fig. \ref{fig:results} cover models learned from a broad range of dataset set-ups with different levels of annotation efforts.
It includes unsupervised training data (Unicorn with Pascal3D+ images), StyleGAN generated data which adds a minute longer annotation effort (our method), a much more expensive data with real images and with key-point annotations (Dib-r on labeled Pascal3D+ images), and a synthetic data (Pixelnerf with ShapeNet dataset) with perfect annotations.

Lastly, we show the final results of our method on bird and horse classes in Fig. \ref{fig:bird}.
3D models of these categories without textures are also shown in Fig. \ref{fig:reb-3d-mdoel}.
Our method achieves realistic 3D predictions for these classes as well even though StyleGAN generated datasets have inconsistencies across views.

\begin{figure}
    \centering
    \includegraphics[width=1\linewidth]{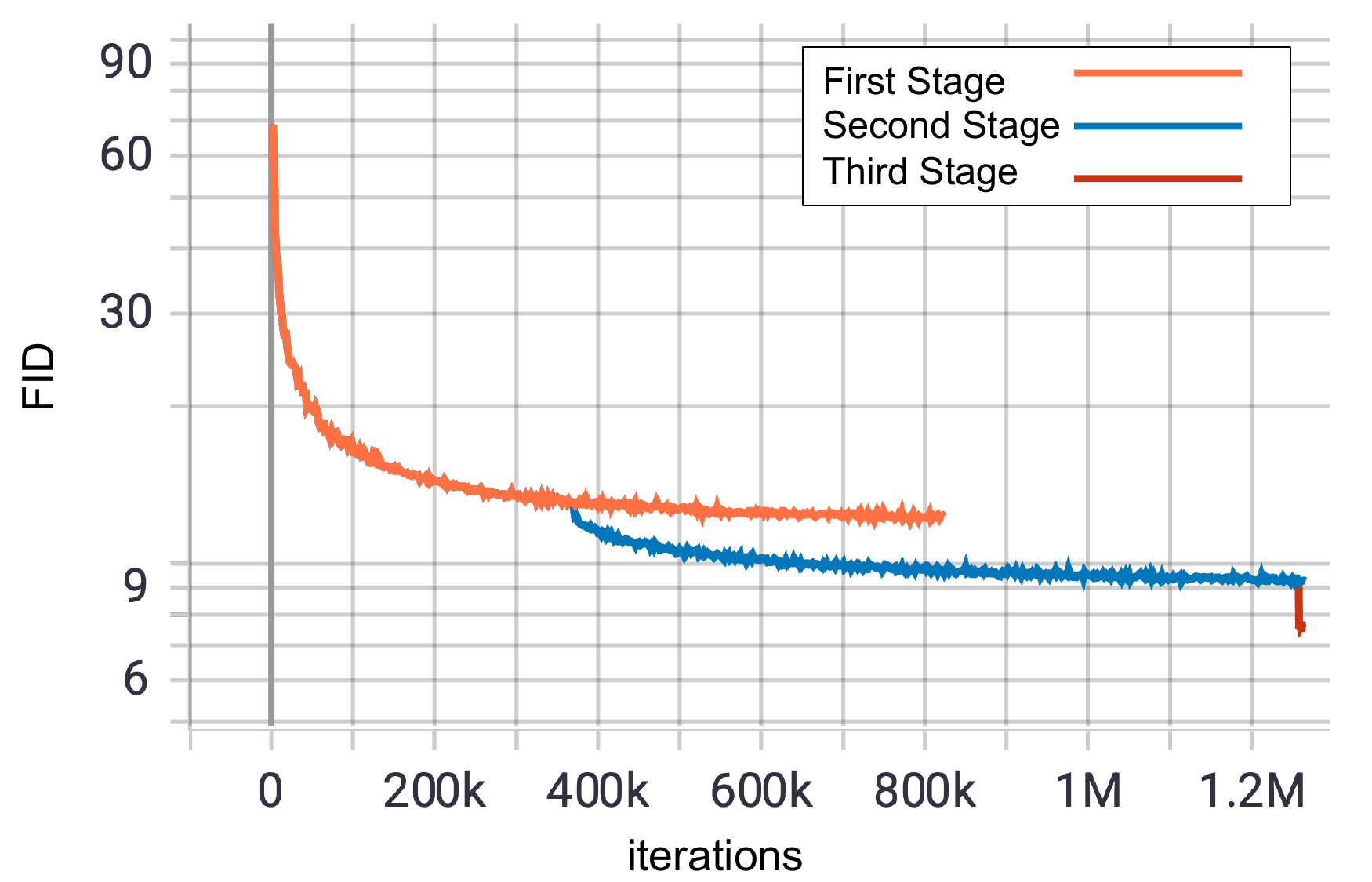}
    \caption{Validation FID curve on Car class with respect to number of iterations.
    As shown, the improvements are not coming from longer iterations but from the progressive training.}
    \label{fig:curve}
\end{figure}

\begin{table*}[t]
\centering
\caption{Ablation study showing results without multi-stage pipeline, without multi-view discriminator, and without the gaussian sampling in the generator on Car dataset. Results are given for same and novel views and contains scores, LPIPS, MSE, SSIM, and 2D mIOU accuracy predictions and GT.}
\begin{tabular}{|l|l|l|l|l|l|l|l|l|l|l|}
\hline
& \multicolumn{5}{c}{Same View} & \multicolumn{5}{|c|}{Novel View}\\
\hline
\textbf{Method} &  FID  $\Downarrow$ & LPIPS  $\Downarrow$ & MSE   $\Downarrow$ & SSIM  $\Uparrow$ & IoU $\Uparrow$ & FID $\Downarrow$ &  LPIPS  $\Downarrow$ & MSE  $\Downarrow$  & SSIM$\Uparrow$  &  IoU $\Uparrow$ \\
\hline
 Ours - Stage I & 10.75 & 0.1011 & 0.0060 & 0.8695 &  0.94 & 11.98  & 0.1091 & 0.0074 & 0.8582 & 0.92\\
 Ours -Stage II & 6.24 &  0.0737 & 0.0039 &0.9027&  0.94 & 9.05 & 0.1012 & 0.0072 & 0.8651 & 0.93 \\
Ours - Stage III & 4.56 & {0.0696} & 0.0040 & 0.9039 & {0.94} &  {6.92} &  {0.0965} & 0.0076 &  0.8632 & {0.93}\\
\hline
No Multi-Stage &  66.59 &  0.1424 &0.0118& 0.8276& 0.91 & 66.71&  0.1485 & 0.0136 & 0.8200 & 0.90\\
Same-View Training & 4.88 & 0.0759 & 0.0037 & 0.9034 & 0.96 & 42.24 & 0.1256 & 0.0105 & 0.8469 & 0.88 \\
Multi-View Training &  6.60 &  0.0741 & 0.0039 &  0.9026 & 0.94 & 10.00 & 0.1017 & 0.0071 & 0.8657 &  0.94 \\
\hline 
No Multi-View disc. &  4.65 &  0.0702 & 0.0039 &  0.9042 & 0.94 & 7.02 & 0.0965 & 0.0076 & 0.8633 &  0.93\\
No Gaussian samp. & 4.67 & 0.0696 &  0.0039 & 0.9043 & 0.94 & 7.05 & 0.0986 &  0.0075 & 0.8625 & 0.93\\
\hline
\end{tabular}
\label{table:abl_res}
\end{table*}

\begin{figure}
    \centering
    \includegraphics[width=1\linewidth]{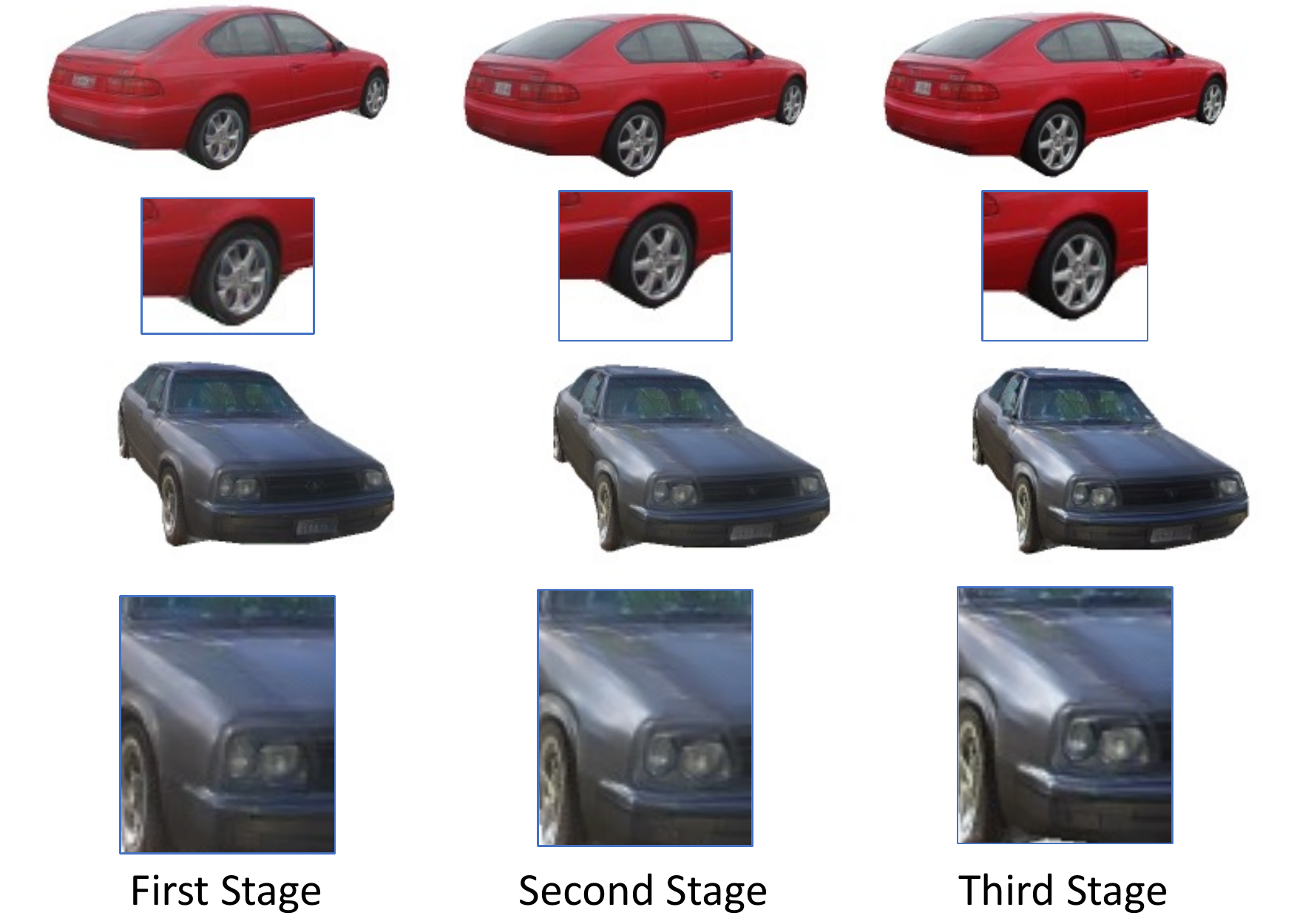}
    \caption{Qualitative results of our models from first, second, and third stage trainings on Car class.
    At each stage, results improve significantly. For example, zooming into tires in the first stage, we see duplicated features. The second stage solves that problem mostly but results are not as sharp as the third stage results. }
    \label{fig:ablation}
\end{figure}

\textbf{Ablation Study.}
First, we analyze the role of each stage in our training pipeline. As given in Table ~\ref{table:results_all}, additional training at each stage improves metrics consistently, especially FIDs and LPIPS, the metrics that are shown to closely correlate with human perception. 
In Fig. \ref{fig:curve}, we provide the training curves of each stage. As can be seen from figure, the improvements are not coming from longer trainings but from the progressive learning.
We provide qualitative comparisons of each stage's output renderings in Fig. \ref{fig:ablation}.
First stage rendering outputs results with a reliable geometry. However, the texture is not realistic, especially tires have duplicated features which is understandable given that the model is minimizing the reconstruction loss from inconsistent multi-view images.
In the second-stage, texture improves significantly over first-stage. Finally, with GAN training in the last stage, the colors look more realistic, sharp, and fine-details are generated.  


We provide additional ablation study in Table \ref{table:abl_res} conducted on Car dataset.
We also provide our each stage scores in the first block to compare the results easily. 
In the second block, we first experiment with no multi-stage training (No Multi-Stage).
This set-up refers to training the model from scratch with the final proposed loss function.
This training results in poor results in all metrics and even worse than our first-stage training results.
Especially, adversarial training  makes the training less stable when the model did not yet learn reliable predictions.
It shows the importance of our multi-stage training pipeline.

Next, we train with Same-view training objectives.
This set-up is used when multi-view images are not available and models have to be trained on single-view images.
We use a similar perceptual objective to Eq. \ref{eq:sv_perc} but with the ground-truth silhouettes as given in Eq. \ref{eq:sv_perc2}.
We also add the silhouette loss from the same view to guide the geometry predictions.
Here the input and target images share the same camera parameters.

\begin{equation}
\label{eq:sv_perc2}
\small
    \mathcal{L}_{p-sv-sil} = ||\Phi_{j}(I^g_{v1}*S^g_{v1}) - \Phi_j(I^r_{v1}*S^g_{v1})||
\end{equation}

\begin{equation}
\label{eq:sil-sv}
\vspace{-1pt}
\small
\mathcal{L}_{sil-sv} = 1 - \frac{||S^g_{v1}\odot S^r_{v1} ||_1}{||S^g_{v1}+ S^r_{v1} - S^g_{v1}\odot S^r_{v1} ||} 
\end{equation}

\begin{equation}
\begin{multlined}
\label{eq:sameview}
\small
\mathcal{L}_{sv} = 
\lambda_{pn} \mathcal{L}_{p-sv-sil} + \lambda_s \mathcal{L}_{sil-sv}  +  \lambda_{lap} \mathcal{L}_{lap}   
\end{multlined}
\end{equation}

The same-view training (model trained with objective from Eq. \ref{eq:sameview}) results are given  in Table \ref{table:abl_res}.
With this set-up, same view results look good quantitatively because the network learns to reconstruct the input view. 
Especially IoU of the same view is better than the other set-ups because the model learns the missing parts of input images and their corresponding silhouettes and make similar predictions on the validation dataset that match the ground-truth.
On the other hand, IoU of novel view score is the worst among all set-ups.
Models trained with single-view objectives struggle generating realistic novel views as can be seen in FIDs, 42.24 novel view FID versus 4.88 same view FID.

We also experiment with multi-view training set-up. This refers to training the model from scratch with the second stage objective.
We compare those results with our second-stage training results. 
We see that better results are achieved with the progressive learning.

In the last block, we run experiments where we train a  discriminator without the multi-view conditioning.
For this experiment, we only update the stage three training.
The discriminator only has the main pipeline without the projection based other view conditioning.
This also results in worse results than our proposed multi-view conditional discriminator, especially in FIDs. 
Since the  multi-view conditional discriminator receives guidance from a given view, it propagates better signals to the generator.

We also experiment the setting which does not have  the sampling from normal distribution.
This change converts the setting to a deterministic model. We observe that sampling provides with a slight diversity in the colors and improves the metrics slightly so we decide to keep it.
The diversity results are given in Fig. \ref{fig:diversity}.
The model does not achieve visible diversity, however, when we take the difference of two images, they are slightly different.
The improvements are not significant but consistent across all metrics.




\newcommand{\interpfigc}[1]{\includegraphics[trim=0 0 0cm 0, clip, height=3.2cm]{#1}}

\begin{figure}
\centering
\scalebox{0.71}{
\addtolength{\tabcolsep}{-5pt}   
\begin{tabular}{ccc}
\interpfigc{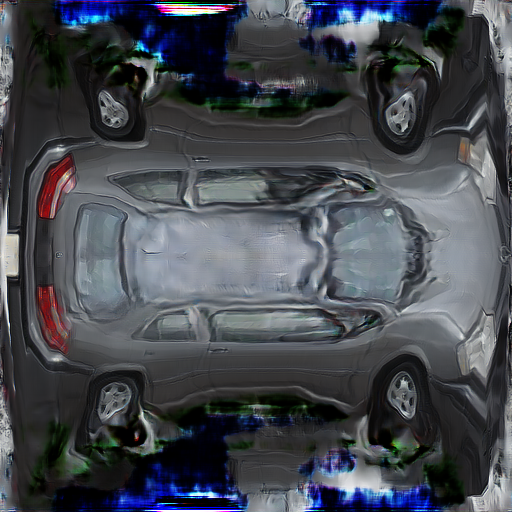} &
\interpfigc{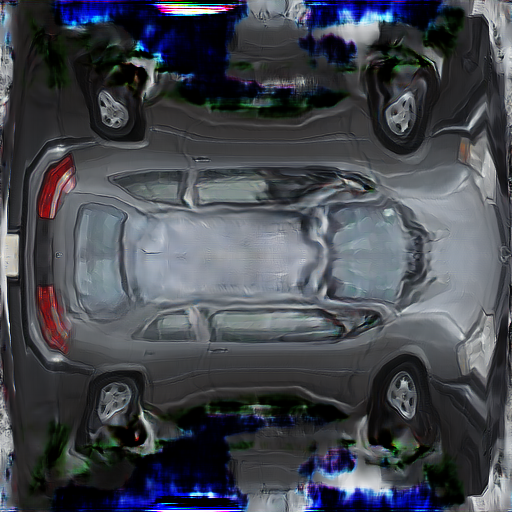} &
\interpfigc{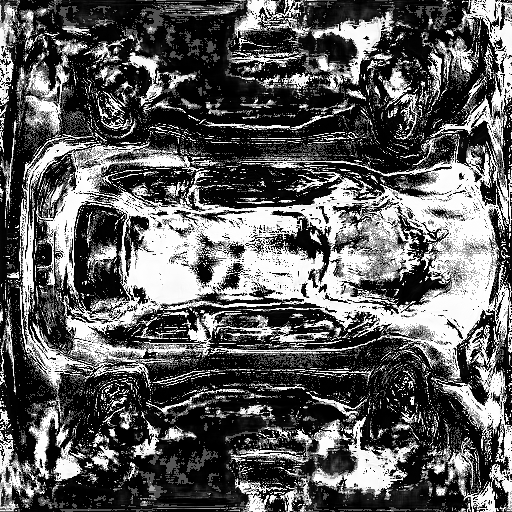} \\
Sample 1 & Sample 2 & Difference Map \\
\interpfigc{Figures/diversity/0_pred.png} &
\interpfigc{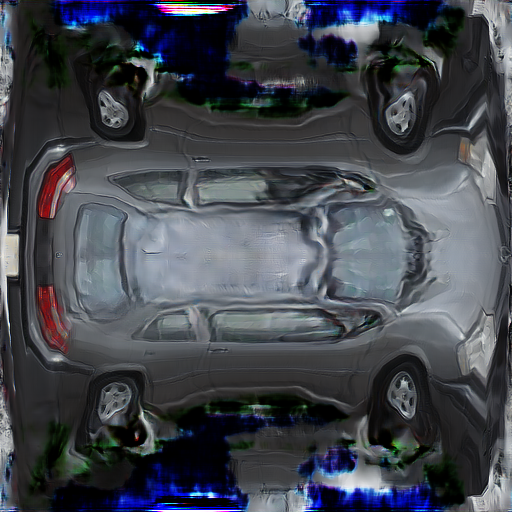} &
\interpfigc{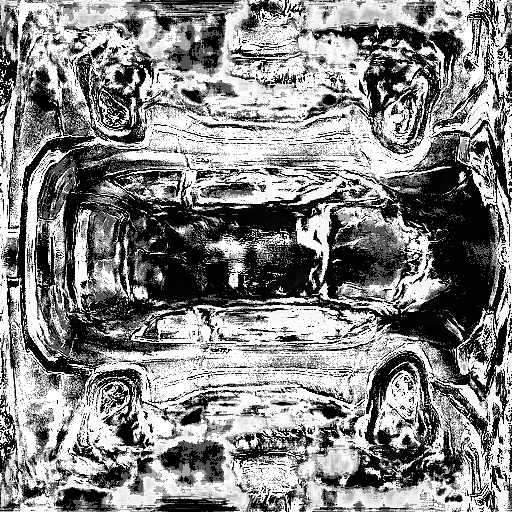} \\
Sample 1 & Sample 3 & Difference Map \\
\interpfigc{Figures/diversity/100_pred.png} &
\interpfigc{Figures/diversity/200_pred.png} &
\interpfigc{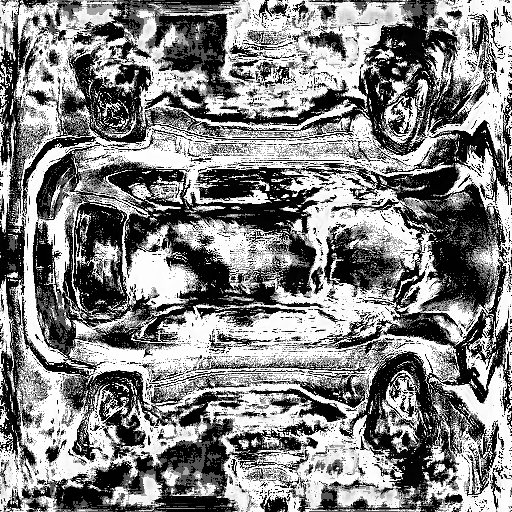} \\
Sample 2 & Sample 3 & Difference Map \\
\end{tabular}}
\caption{Texture predictions with different sampled codes and with same input image. The model does not achieve visible diversity. However, when we take the difference of two images, they are slightly different.}
\label{fig:diversity}
\end{figure}

\section{Conclusion}

In this work, we present a method to reconstruct high-quality textured 3D models from single images.
Our method learns from GAN generated images and bypasses the reliance on labeled multi-view datasets or expensive 3D scans.
GAN generated dataset is labeled in mass and requires a total of 1 minute human-labor. 
Because GAN generated dataset is noisy and not strictly consistent across views, we propose  a novel multi-stage training pipeline and adversarial training set-up.
We achieve significant improvements over previous methods whether they were trained on GAN generated images or on real images.

\textbf{Limitations.} First limitation of our work is that we deform our final meshes from a sphere and cannot handle objects with holes similar to previous works that build their work on mesh representations obtained by deforming from spheres \cite{kanazawa2018learning, zhang2020image, pavllo2020convolutional}.
Another limitation we observe is the different 3D model qualities  we obtain across different categories. Specifically, our generated 3D models are better quality for the car class than the bird class.
We also observe the same on the StyleGAN image generations results between car and bird classes. 
One reason for that is that StyleGAN model is trained on 5.7M car images whereas for bird category it is only trained on 48k bird images. 
We acknowledge that the performance of our model is correlated with the performance of the StyleGAN model it learns from. 
Even though, our model does not need annotated images, StyleGAN model requires a large amount of unlabeled data. Learning GAN models on limited data is an important future direction for this work \cite{karras2020training}.

{
\bibliographystyle{ieee}
\bibliography{egbib}
}

\end{document}